%% file: acl2023.tex
\pdfoutput=1

\documentclass[11pt]{article}

\usepackage[final]{ACL2023}

\usepackage{times}
\usepackage{latexsym}
\usepackage{graphicx}
\usepackage{hyperref}
\usepackage{fancyvrb} 
\usepackage{tcolorbox} 
\usepackage{booktabs}
\usepackage{svg}

\usepackage[utf8]{inputenc}
\usepackage{enumitem}

\usepackage{natbib}

\usepackage[T1]{fontenc}

\usepackage[utf8]{inputenc}

\usepackage{microtype}

\usepackage{inconsolata}

\title{Ta-G-T: Subjectivity Capture in Table to Text Generation via RDF Graphs}

\author{
  Ronak Upasham \quad
  Tathagata Dey \quad
  Pushpak Bhattacharyya \\
  Department of Computer Science and Engineering \\
  Indian Institute of Technology Bombay \\
  ronakupasham@cse.iitb.ac.in
}

\begin{document}
\maketitle
\begin{abstract}
In Table-to-Text (T2T) generation, existing approaches predominantly focus on providing objective descriptions of tabular data. However, generating text that incorporates subjectivity, where subjectivity refers to interpretations beyond raw numerical data, remains underexplored. To address this, we introduce a novel pipeline that leverages intermediate representations to generate both objective and subjective text from tables. Our three-stage pipeline consists of: 1) extraction of Resource Description Framework (RDF) triples, 2) aggregation of text into coherent narratives, and 3) infusion of subjectivity to enrich the generated text. By incorporating RDFs, our approach enhances factual accuracy while maintaining interpretability. Unlike large language models (LLMs) such as GPT-3.5, Mistral-7B, and Llama-2, our pipeline employs smaller, fine-tuned T5 models while achieving comparable performance to GPT-3.5 and outperforming Mistral-7B and Llama-2 in several metrics. We evaluate our approach through quantitative and qualitative analyses, demonstrating its effectiveness in balancing factual accuracy with subjective interpretation. To the best of our knowledge, this is the first work to propose a structured pipeline for T2T generation that integrates intermediate representations to enhance both factual correctness and subjectivity.
\end{abstract}

\begin{figure}[!ht]
    \centering
    \includegraphics[width=\columnwidth]{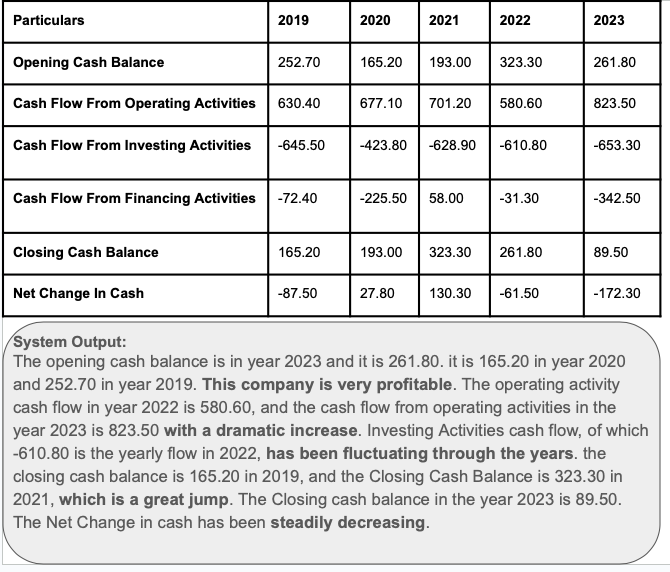}
    \setlength{\belowcaptionskip}{-16pt}
    \caption{\textbf{Generating text with subjectivity from a table}: The table contains the Cash Flow Statement of a company over 5 years. The text below is generated from our pipeline which describes the tabular information where the bold phrases refer to the infused subjectivity.}
    \label{fig:subjobj}
\end{figure}

\section{Introduction}

Table-to-text (T2T) generation converts structured data into natural language summaries. While most prior work focuses on generating objective descriptions, we introduce a pipeline that incorporates contextual subjectivity—interpretive insights grounded in data trends. These subjective elements do not involve personal opinions but rather evaluative statements that enhance understanding.

For example, a table showing revenue may yield an objective sentence like \emph{"The company generated \$500,000 last year."} In contrast, a subjective version might say, \emph{"Despite generating \$500,000, the company's growth has slowed compared to the previous year."} Such interpretations remain fact-based but provide analytical context, unlike hallucinations that introduce unsupported content.

We propose Ta-G-T (Table to RDF Graph to Text), a novel three-stage pipeline consisting of: (1) RDF triple extraction to structure tabular data and preserve factual accuracy; (2) sentence aggregation for fluent narrative construction; and (3) subjectivity infusion for adding contextual interpretations.

This modular design offers several advantages: each stage can be independently optimized and analyzed, enabling better control over output quality and facilitating granular error diagnosis. By using RDF as an intermediate representation, we ground generation in structured semantics and reduce hallucinations. The pipeline also supports reusability across domains and can incorporate additional capabilities—such as domain-specific styles or reasoning—without redesign.

Our approach also provides a computationally efficient alternative to LLMs. While models like GPT-3.5 offer generalization, they often underperform on structured inputs. Using smaller fine-tuned models (e.g., T5-large), we achieve strong performance on both factual and subjective dimensions.

\begin{figure}[!ht]
    \centering
    \includegraphics[width=\columnwidth]{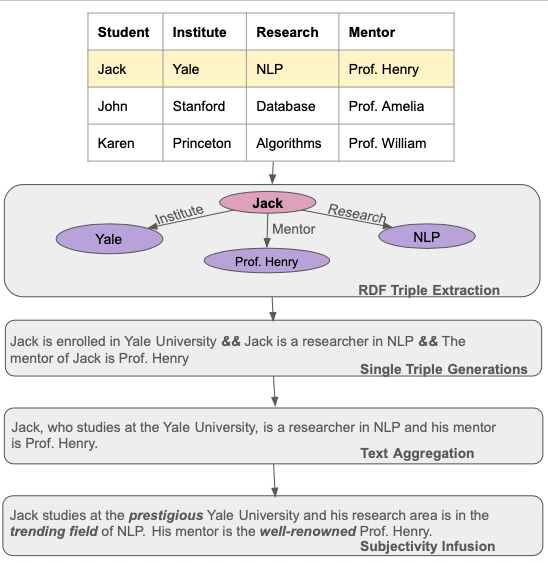}
    \setlength{\belowcaptionskip}{-16pt}
    \caption{\textbf{Pipeline for subjective text generation from tables}: Illustration of the stages of our proposed pipeline.}
    \label{fig:pipeline}
\end{figure}

Experiments demonstrate that our approach rivals GPT-3.5 and outperforms Mistral-7B and Llama-2 in both objective and subjective quality. Figure~\ref{fig:pipeline} illustrates our method, using a non-numeric table with example highlights to show each transformation stage.

\section{Problem Statement}
\label{probst}

We define subjectivity in T2T generation as contextual interpretation of structured data, where insights, trends, and evaluative phrases are added without deviating from factual content.

\textbf{Input:} \textit{Tabular data structured in rows and columns.}

\textbf{Output:} \textit{Fluent, subjective text that both describes and interprets the data.}

For example, in a table showing a company’s cash flow, saying \emph{"dramatic increase"} for a rise from 580.60 to 823.50 adds a relative judgment grounded in the data. Similarly, \emph{"fluctuating over the years"} conveys trend insight rather than listing values.

Subjectivity is context-sensitive. A 5\% profit margin might be \emph{"healthy"} in a stagnant market but \emph{"underwhelming"} in a growth sector. A 5°C drop could be \emph{"a cool spell"} or \emph{"a minor dip"} depending on seasonal expectations.

Our modular pipeline enables these subjective interpretations to be introduced systematically, ensuring factual grounding through RDF and interpretability through sentence-level generation and aggregation. Moreover, the pipeline design supports audience-aware customization: an investor report might emphasize growth (\emph{"strong financial performance"}), whereas a stakeholder summary might prioritize stability (\emph{"consistent cash flow trends"}).

\subsection{Contributions}

In this paper, we present a novel pipeline for generating factually correct and subjective text from tables. Our contributions are:

\begin{enumerate}

    \item Proposal of a \textbf{novel three-stage pipeline} that extracts RDF triples from tables, generates text from these triples, aggregates the text into coherent sentences, and infuses subjectivity to provide nuanced interpretations. To the \textit{best of our knowledge}, we are the first to test a pipelined architecture to capture both factual accuracy and subjectivity from tables (see Section~\ref{implementation}).

    \item Development of a metric for \textbf{automatic evaluation of subjectivity}, leveraging a fine-tuned T5-large model on the Movies Subjectivity Dataset. This metric quantifies the percentage of subjective statements in generated text, providing a novel approach to measuring subjectivity beyond human evaluation (see Section~\ref{quantAnalysis}).

    \item Achievement of \textbf{improvements in performance}, with a METEOR score of 25.46\% and a BERTScore of 82.50\%. Human evaluations further demonstrated that the generated text was more coherent and fluent compared to baseline T5 models (see Section~\ref{results}).
\end{enumerate}

\subsection{Motivation}

Structured data is abundant in domains like finance, healthcare, governance, and sports—but it often lacks interpretability in raw form. Subjective interpretation bridges this gap by highlighting trends, comparisons, or anomalies that aid decision-making.

For instance, in business reporting, stating that dividends are “steady” or that revenue “declined slightly” provides readers with actionable insights. In healthcare, interpreting patient metrics as “improved significantly” or “showing minor decline” can support better clinical decisions. These interpretations require controlled subjectivity—expressing insights without hallucination. Our pipeline addresses this by inserting subjectivity in the final stage, after factual correctness has been ensured.

Additionally, the modular structure enhances adaptability: components can be reused or upgraded (e.g., adapting subjectivity for specific audiences or domains), making the approach extensible and robust across real-world use cases.

\section{Related Works}
\label{related}

Table-to-text (T2T) generation has been advanced by Pre-trained Language Models (PLMs) like BERT \cite{devlin-etal-2019-bert}, T5 \cite{t5}, and GPT \cite{gpt}, which have proven effective in handling structured inputs. Models such as TAPAS \cite{herzig2020tapas} and TaBERT \cite{yin2020tabert} integrate table structures into PLMs, while T5 and BART have been shown to outperform graph-based approaches using linearized inputs \cite{kale2020text, ribeiro2020investigating}.

Earlier pipeline-based approaches decomposed D2T tasks into stages like content selection and surface realization \cite{gatt2018survey, reiter1997building, reiter-2007-architecture}. Neural pipelines \cite{ferreira2019neural} and zero-shot frameworks \cite{laha2020scalable} demonstrated the value of modular generation, while \cite{shen2020neural} addressed hallucinations through alignment techniques.

Intermediate representations like logical forms have been explored to enhance coherence \cite{logic2text, plog}, but these works largely focus on objective reporting.

Datasets such as ToTTo \cite{tottodata}, Rotowire \cite{rotowire}, and LogicNLG \cite{logicnlg} have supported T2T research. While ToTTo and Rotowire target surface-level factual reporting, LogicNLG emphasizes logical inference, with limited focus on subjective interpretation.

Graph-to-text generation is also a growing area. GPT-3 has been evaluated for this task in zero-shot settings \cite{yuan2023evaluating}, though challenges in grounding remain. PlanGTG \cite{he2025evaluating} improves LLM performance via planning annotations. A taxonomy of Graph2text and Graph2token strategies is proposed in \cite{yu2025graph2text}, while \cite{wang2021stage} enhances graph encoding through tree-level embeddings. These studies deal with structured text generation but do not address subjectivity in graph-derived text.

\section{Dataset}

\textbf{Ta2TS Dataset:} We utilize the Ta2TS Dataset (Table-to-Text with Subjectivity) \cite{ta2ts}, as the primary evaluation dataset for our pipeline. This dataset is specifically designed for generating both objective and subjective text from tables, making it an important resource for assessing the effectiveness of our method. The Ta2TS dataset includes three types of tables: financial statements, weather forecasts, and sports data. These tables were originally collected from diverse online sources such as \emph{Groww, the Indian Meteorological Department, ESPN Cric Info, IPL, and Goal}. After filtering, the final dataset comprises 3,849 tables, distributed across these domains. This balanced distribution is detailed in Appendix \ref{appendix:ta2ts}. The Ta2TS dataset serves as the benchmark for evaluating the ability of our pipeline to generate factually accurate and subjectively rich descriptions.

\textbf{WebNLG Dataset:} \cite{webnlg} We employ the WebNLG dataset to train the first stage of our pipeline, which converts structured data into natural language descriptions. This dataset consists of RDF (Resource Description Framework) triples extracted from DBpedia \cite{dbpedia}, a large-scale structured knowledge base derived from Wikipedia. RDF triples are representations of factual knowledge, where each triple consists of a subject, predicate, and object. For example:

\begin{itemize}[noitemsep]
    \item \textbf{Subject:} Barack Obama
    \item \textbf{Predicate:} born in
    \item \textbf{Object:} Honolulu
\end{itemize}

This triple conveys the fact that "Barack Obama was born in Honolulu." The WebNLG dataset pairs such triples with corresponding human-written natural language descriptions across multiple domains, including sports, politics, and entertainment. We fine-tune a T5-large model on WebNLG to generate textual descriptions for individual RDF triples extracted from tables. This step ensures the factual accuracy of the generated text before further aggregation and subjectivity infusion.

\textbf{Wiki Neutrality Corpus (WNC):} We use the Wiki Neutrality Corpus (WNC) \cite{pryzant} to train the third stage of our pipeline, which transforms neutral aggregated text into subjective narratives. WNC consists of paired sentences, where one version is neutral (edited for encyclopedic tone) and the other retains subjective phrasing. These sentence pairs originate from Wikipedia edits, providing a reliable training resource for modeling subjectivity infusion. In our approach, we fine-tune a T5-large model on the reversed WNC dataset to inject evaluative and subjective expressions into the text, allowing for more interpretative and expressive descriptions of tabular data.

\section{Implementation}
\label{implementation}
Our proposed pipeline for table-to-text generation introduces a novel three-stage framework that leverages intermediate RDF representations, sentence aggregation, and subjectivity infusion. This structured design, which has not been explored before in table-to-text generation, enhances factual accuracy, fluency, and interpretability while allowing independent optimization of each component.

\subsection{RDF Extraction and Single Triple Generation}  
The first stage of our pipeline introduces RDF triples as an intermediate representation for T2T generation. Unlike direct table-to-text approaches, we first convert tabular data into structured RDF triples to enhance factual accuracy. The first column serves as the subject, column headers as predicates, and cell values as objects, forming a single-star RDF graph per row.

This transformation follows a deterministic heuristic, ensuring a direct, error-free mapping from tables to RDF triples. Since no probabilistic decisions are involved, this stage requires no separate accuracy evaluation, as every table entry is systematically converted without modification. This structured representation provides a reliable foundation for later text generation.

To generate textual descriptions from RDF triples, we fine-tune a T5-large model on the WebNLG dataset. While WebNLG does not cover all domains in Ta2TS, its diverse RDF triples across domains like \emph{Sports Team, University, and Written Work} provide strong structural mappings that generalize well. The dataset's high-quality human-annotated references enable robust RDF-to-text learning, allowing the model to effectively transfer sentence structures and relationships to unseen domains such as finance and weather.

\subsection{Sentence Aggregation}
This stage introduces another novel contribution: text aggregation for table-to-text generation. While previous methods have focused on single-step text generation, our pipeline explicitly incorporates sentence aggregation to improve coherence and fluency. 

To train our aggregation model, we construct a synthetic dataset by leveraging the first-stage model. This model is used to generate independent sentences for RDF triples, which serve as input to the synthetic dataset. The output column contains corresponding aggregated, coherent text from the WebNLG dataset. For instance, an input sequence like:  
\emph{"John is a boy. | John is smart. | John studies in school."}  
corresponds to an output of:  
\emph{"John is a smart boy who is studying in a school."}

By training a T5-large model on this dataset, we enable it to merge semantically related sentences, resolve co-references, and introduce conjunctions. This structured aggregation eliminates redundancy and produces fluid, logically structured narratives that maintain the integrity of tabular data.

\subsection{Subjectivity Infusion}
The third stage of our pipeline introduces the novel technique of style transfer for objective-to-subjective transformation. Instead of simply generating descriptions, this step enhances interpretability by incorporating evaluative language. 

We fine-tune a T5-large model on the reverse of the WNC, where the original subjective text is treated as the target output. This model learns to introduce adjectives, expressive phrases, and subjective commentary while preserving the factual correctness of the content. The transformation enhances readability and engagement, allowing the generated text to convey human-like interpretations rather than just raw numerical descriptions.

\subsection{Advantages of a Modular Pipeline Design}
The modular design of the pipeline offers several advantages, combining structured functionality, interpretability, and adaptability. By dividing the process into stages—RDF extraction, text aggregation, and subjectivity infusion—the pipeline allows for independent optimization of components, giving flexibility and scalability.

A key innovation is the use of RDF triples as intermediate representations, which enhances factual accuracy by grounding the text in structured data and reducing hallucinations, as demonstrated by \cite{kgHallucination}. This structured approach also improves error analysis, as issues like factual inaccuracies or poor fluency can be traced to specific stages.

The modular framework supports reusability, where components like the subjectivity infusion stage can be applied across tasks or datasets, and scalability, allowing the addition of new functionalities such as analytical reasoning without re-designing the system. Using smaller models like T5 large provides a computationally efficient alternative to LLMs, achieving strong performance on structured data while balancing both objective and subjective elements.

\subsection{Baselines}
\label{baselines}

We compare our pipeline against two baselines from \cite{ta2ts}: fine-tuned T5 models and prompted LLMs, and directly use their results. For the first, T5-base, T5-large, and T5-3b were fine-tuned on linearized tables, with varying degrees of contextual prefixes to assess the impact of metadata and row structure. The second baseline uses LLMs such as GPT-3.5, Mistral-7b \cite{mistral}, and Llama-2 \cite{llama} in 0-shot, 1-shot, and few-shot prompting setups. While capable of generating fluent outputs, these models often lack factual consistency without task-specific fine-tuning.

\subsection{Training Details}
We fine-tuned the T5-large models for each of the three stages of the Ta-G-T pipeline for 3, 50, and 50 epochs, respectively. The number of epochs was determined through careful monitoring of the loss and performance metrics across the training process. A learning rate of \(2 \times 10^{-5}\) was utilized with a batch size of 4. On average, fine-tuning the models required approximately 1, 12 and 12 hours respectively, for each stage. All experiments were conducted using an NVIDIA A100 GPU with 80GB of memory. We implemented the Adafactor optimizer with a learning rate of \(1 \times 10^{-3}\) and a decay rate of 0.8.

\section{Results}
\label{results}

We evaluated the generated texts of both fine-tuned models and LLMs through automatic and human evaluation methods. Below, we present the automatic metrics and analysis of our pipeline compared to baseline models.

\subsection{Quantitative Analysis}
\label{quantAnalysis}

\begin{table*}[!ht]
    \centering
    \begin{tabular}{ r c c c c c }
        \hline
        \textbf{Model} & \textbf{Bleu-4} & \textbf{Meteor} & \textbf{Rouge-L} & \textbf{BertScore}\\
        \hline
        T5-base, w/o-pf & 2.89 & 22.17 & 21.11 & 80.43 \\
        T5-base, pf-w/o-ct & 3.22 & 24.60 & 22.97 & 84.14\\
        T5-base, pf-ct & \textbf{3.25} & 24.98 & 23.45 & \textbf{85.15} \\
        \hline
        T5-large, w/o-pf & 2.81 & 23.54 & 21.98 & 81.43\\
        T5-large, pf-w/o-ct & 2.94 & 23.65 & 22.31 & 82.22 \\
        T5-large, pf-ct & \underline{3.20} & 24.11 & \textbf{23.67} & 84.37 \\
        \hline
        T5-3b, w/o-pf & 2.57 & 20.11 & 21.39 & 81.58 \\
        T5-3b, pf-w/o-ct & 2.63 & 21.77 & 22.64 & 82.00\\
        T5-3b, pf-ct & 2.67 & 22.03 & 22.78 & 82.28\\
        \hline
        Llama-2, 0-shot & 2.17 & 17.44 & 19.32 & 56.71\\
        Llama-2, 1-shot & 2.31 & 18.13 & 19.45 & 59.00\\
        Llama-2, 5-shot & 2.84 & 19.65 & 19.88 & 63.14\\
        \hline
        Mistral-7b, 0-shot & 2.21 & 18.09 & 13.65 & 30.01 \\
        Mistral-7b, 1-shot & 2.48 & 18.76 & 18.11 & 40.83 \\
        Mistral-7b, 3-shot & 2.69 & 19.26 & 18.76 & 29.64\\
        Mistral-7b, 7-shot & 2.08 & 20.17 & 13.92 & 29.87 \\
        \hline
        GPT3.5, 0-shot & 2.64 & 25.3 & 21.9 & 81.70 \\
        GPT3.5, 1-shot & 3.08 & \textbf{26.28} & 23.02 & 83.87 \\
        GPT3.5, 3-shot & 2.98 & \underline{25.97} & \underline{23.57} & \underline{84.78} \\
        GPT3.5, 7-shot & 2.48 & 25.29 & 23.24 & 84.42 \\
        \hline
        \textbf{Ta-G-T} & 1.63 & 25.46 & 18.57 & 82.50 \\
        \hline
    \end{tabular}
    \setlength{\belowcaptionskip}{-11pt}
    \caption{\textbf{Metrics Comparison}: Results for Bleu-4, Meteor, Rouge-L, and BERTScore across models. The highest scores are in bold, and the second highest in underline. Scores are out of 100.}
    \label{tab:quanti}
\end{table*}

Evaluating subjectivity in text generation presents unique challenges due to its nuanced, context-dependent nature. Traditional metrics such as BLEU-4, METEOR, ROUGE-L, and BERTScore remain the standard for assessing objective text generation but are insufficient for capturing subjective nuances like tone, evaluative language, or interpretive insight. These metrics focus primarily on lexical overlap and semantic similarity, which do not reflect subjective or contextual richness. For example, a phrase like \emph{"steady improvement"} may carry subjective interpretation, yet still share low lexical overlap with a reference phrase like \emph{"gradually rising numbers."}

To address this limitation, we introduce an \textbf{automatic subjectivity evaluation} method. This involves fine-tuning a T5-large model on the Movies Subjectivity Dataset by \cite{pangLee}, which contains 5,000 subjective movie review sentences and 5,000 objective plot summary sentences. The model is trained for binary classification (subjective vs. objective) and achieves 98\% accuracy on the test set. This classifier is applied sentence-wise to the generated outputs, and the percentage of sentences classified as subjective is used as a quantitative subjectivity score.

Applying this classifier on outputs from the Ta2TS dataset, we obtain subjectivity scores of \textbf{24.62\%} for Weather, \textbf{12.35\%} for Sports, and \textbf{14.52\%} for Finance. These indicate the presence of meaningful subjective elements in the generated text, and are particularly informative in ablation studies (see Section~\ref{ablation}) where we compare generations with and without the subjectivity infusion stage. The significant drop in subjectivity when the third stage is removed confirms the contribution of this component.

In addition to subjectivity evaluation, Table~\ref{tab:quanti} presents BLEU-4, METEOR, ROUGE-L, and BERTScore for various baselines and our pipeline. Notably, \textbf{Ta-G-T outperforms both Mistral-7b and Llama-2 across all four metrics}, confirming its superiority over these LLMs in generating structured, accurate, and coherent text from tables. Interestingly, all T5 variants, including T5-base and T5-large (fine-tuned on the Ta2TS dataset), also outperform Mistral and Llama. This suggests that smaller but task-specific fine-tuned models may generalize better to structured inputs like tables, while open-domain LLMs struggle with alignment and formatting in data-to-text scenarios.

While \textbf{GPT-3.5 consistently achieves the highest scores across all metrics}, it is important to note that this is expected given its scale and extensive pretraining. However, even in comparison to GPT-3.5, \textbf{Ta-G-T demonstrates competitive performance}, especially in METEOR (25.46 vs. 26.28) and BERTScore (82.50 vs. 84.78), despite being significantly smaller and more efficient.

Crucially, \textbf{Ta-G-T was not fine-tuned on the Ta2TS dataset}, in contrast to all T5 baselines, which were directly trained on Ta2TS. The pipeline is instead trained only on intermediate stages like WebNLG and WNC and evaluated directly on Ta2TS without adaptation. That it still performs close to or better than some of the fine-tuned baselines in METEOR and BERTScore highlights its generalizability and strong architectural design. These findings reinforce the value of our structured, modular pipeline, which balances factual consistency, fluency, and subjective enrichment—even in the absence of direct supervision on the evaluation dataset.

While traditional metrics remain limited in evaluating subjectivity, our combined use of automatic subjectivity detection and standard metrics provides a holistic framework. Ta-G-T demonstrates a strong balance between factual and subjective generation, and its performance across diverse metrics validates the pipeline's effectiveness for structured data-to-text tasks.

\begin{figure*}[t]
    \centering
    \includegraphics[width=\textwidth]{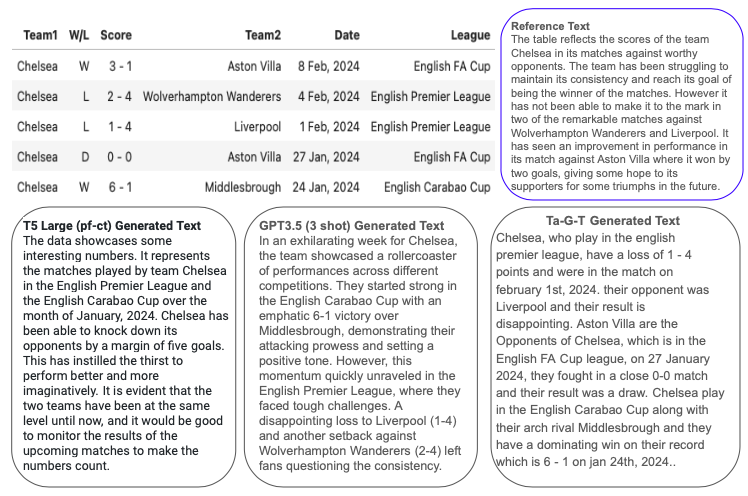}
    \caption{\textbf{Example of texts generated by T5-large (pf-ct), GPT-3.5 (3-shot) and Ta-G-T pipeline}: In the figure, the given table and reference text (ground truth) are taken from the Ta2TS dataset, and at the bottom, the generated texts are shown.}
    \label{fig:gen_ex}
\end{figure*}

\subsection{Qualitative Analysis}
\label{qualitativeanal}

To ensure consistency and reliability in our human evaluation process, we employed three evaluators. These evaluators were provided with a detailed set of scoring guidelines to assess the generated text based on four key parameters: \textbf{coherence}, \textbf{coverage}, \textbf{accuracy}, and \textbf{subjectivity capture}. Each evaluation was performed on a scale of 1 to 10, where 1 represents the lowest quality and 10 represents the highest quality. Detailed instructions to evaluators and their background are provided in Appendix \ref{appendix:humaneval}. For an example table from the Ta2TS dataset, generations from multiple models and the reference text are illustrated in Figure \ref{fig:gen_ex}

\begin{table}[ht]
    \centering
    \begin{tabular}{c c c c c c}
        \hline
        & \textbf{Eval.} &\textbf{Coh.} & \textbf{Cov.} & \textbf{Acc.} & \textbf{Sub.} \\
        \hline
        & \textbf{A} & 8.7 & 9.0 & 8.8 & 9.1\\
        GPT & \textbf{B} & 8.4 & 8.8 & 8.9 & 9.0\\
        & \textbf{C} & 8.9 & 9.0 & 9.1 & 9.1\\
        \hline
        \textbf{H-Mean} && \textbf{8.66} & \textbf{8.93} & \textbf{8.93} & \textbf{9.06}\\
        \hline
        & \textbf{A} & 8.3 & 8.2 & 8.7 & 8.8\\
        T5 & \textbf{B} & 8.1 & 8.0 & 8.5 & 9.0\\
        & \textbf{C} & 8.4 & 8.3 & 8.8 & 9.0\\
        \hline
        \textbf{H-Mean} && \textbf{8.26} & \textbf{8.16} & \textbf{8.66} & \textbf{8.93}\\
        \hline
        & \textbf{A} & 8.4 & 8.7 & 8.9 & 8.7\\
        Ta-G-T & \textbf{B} & 8.4 & 8.7 & 8.9 & 8.6\\
        & \textbf{C} & 8.3 & 8.6 & 8.8 & 8.5\\
        \hline
        \textbf{H-Mean} && \textbf{8.36} & \textbf{8.66} & \textbf{8.86} & \textbf{8.59}\\
        \hline
    \end{tabular}
    \setlength{\belowcaptionskip}{-10pt}
    \caption{\textbf{Human evaluation}: the Harmonic Mean (H-Mean) of all 100 generated samples over the features of Coherence (Coh.), Coverage (Cov.), Accuracy (Acc.) and Subjectivity Capture (Sub.).}
    \label{tab:humanEval}
\end{table}

The human evaluation scores of GPT-3.5, T5-large (pf-ct) and Ta-G-T are presented in Table \ref{tab:humanEval} across the three evaluators (A, B, C). The harmonic mean of all scores is used because it provides a more sensitive distinction between values, particularly when dealing with outliers or extreme values, compared to the arithmetic mean. The details of Inter-Annotator Agreement are provided in Appendix \ref{appendix:iaa}

The results of Ta-G-T show a strong performance in coverage and accuracy, surpassing the T5-large (pf-ct) model. Ta-G-T consistently included more relevant details from the table, ensuring higher factual correctness. This reflects the pipeline's ability to generate text with a broader scope of the table’s data, making it more informative. However, subjectivity capture for Ta-G-T was comparatively weaker. While the pipeline excelled in generating factually accurate and comprehensive summaries, it struggled to incorporate subjective phrases as effectively as T5-large (pf-ct), which exhibited a stronger ability to generate emotion-rich, evaluative language.

The T5-large (pf-ct) model achieved higher subjectivity scores, showing a better balance between factual content and subjective interpretation. However, its lower coverage score indicates that important facts were often omitted, reducing the overall completeness of the generated text. Additionally, the T5 model also faced issues with coherence, primarily due to unresolved co-reference, affecting the logical flow of the narrative. In contrast, GPT-3.5 (3-shot) maintained a consistently high performance across all parameters, with particularly strong subjectivity capture. However, our Ta-G-T pipeline’s higher factual accuracy and coverage make it a valuable alternative for generating factually consistent text, particularly in domains where comprehensive data coverage is crucial. Overall, the results indicate that while Ta-G-T excels in generating comprehensive and accurate summaries, it lags slightly in generating rich subjective content. Nevertheless, the model offers a balanced trade-off between factual accuracy and subjectivity, making it a robust choice for tasks requiring a blend of objectivity and nuanced interpretation.

\subsection{Ablation Study}
\label{ablation}

To evaluate the impact of the subjectivity infusion stage in our pipeline, we conduct an ablation study by comparing the subjectivity levels in the generated text with and without the third stage. The subjectivity of the generated outputs is measured using our automatic subjectivity evaluation metric (detailed in Section~\ref{quantAnalysis}).

Table~\ref{tab:ablation} presents the subjectivity percentages computed across the three domains of the Ta2TS dataset. When the full pipeline is applied, the generated text exhibits significantly higher subjectivity levels compared to the versions without the third stage (WS). These results highlight the effectiveness of the subjectivity infusion stage in enriching the generated narratives with evaluative and interpretative language, moving beyond purely factual descriptions. Furthermore, this ablation study validates that our modular pipeline effectively incorporates subjective elements.

\begin{table}[!ht]
    \centering
    \begin{tabular}{l c c}
        \hline
        \textbf{Domain} & \textbf{Full (\%)} & \textbf{WS (\%)} \\
        \hline
        Finance & 14.52 & 4.32 \\
        Sports & 12.35 & 4.85 \\
        Weather & 24.62 & 3.97 \\
        \hline
    \end{tabular}
    \setlength{\belowcaptionskip}{-10pt}
    \caption{\textbf{Subjectivity Ablation Study:} Subjectivity percentages in text generated with (Full) and without (WS) the subjectivity infusion stage.}
    \label{tab:ablation}
\end{table}

In addition to analyzing the subjectivity stage, we also investigate the impact of the aggregation stage on the overall text quality and coherence. For this, we remove the aggregation stage (Stage 2) and directly feed the outputs of the first stage (RDF triple generation) into the subjectivity infusion model. This configuration allows us to assess whether the aggregation component is necessary for producing fluent and logically structured text.

Evaluation results using BLEU, METEOR, ROUGE-L, and BERTScore metrics reveal a noticeable decline in performance across all domains when the aggregation stage is omitted. Table~\ref{tab:ablationExtra} summarizes these findings. The generated outputs in this setup tend to be fragmented, repetitive, or incoherent, indicating that the lack of intermediate aggregation hinders the model’s ability to generate cohesive narratives. These observations confirm the importance of the aggregation stage in consolidating data into a structured, human-readable format before stylistic transformation. This second ablation thus demonstrates that both the aggregation and subjectivity infusion stages contribute critically to the overall quality and readability of the final generated text.

\begin{table}[!ht]
    \centering
    \begin{tabular}{l c c c}
        \hline
        \textbf{Metric} & \textbf{WA} & \textbf{WS} & \textbf{Ta-G-T}\\
        \hline
        BLEU-4 & 1.00 & 1.19 & 1.63 \\
        METEOR & 21.29 & 22.71 & 25.46 \\
        ROUGE-L & 16.50 & 17.02 & 18.57 \\
        BERTScore & 80.66 & 81.12 & 82.50 \\
        \hline
    \end{tabular}
    \setlength{\belowcaptionskip}{-10pt}
    \caption{\textbf{Ablation Results:} Metric scores for text generations without aggregation stage (WA), and without subjectivity stage (WS).}
    \label{tab:ablationExtra}
\end{table}

\section{Conclusion and Future Work}
In this paper, we introduced a novel pipeline for table-to-text generation that combines subjectivity with factual objectivity, addressing a gap in previous works focused solely on objective text generation. Our multi-stage pipeline leverages RDF triples for triple extraction, sentence aggregation, and subjectivity infusion, demonstrating promising results. While GPT3.5 performed better, our smaller fine-tuned models beat LLMs like Llama-2 and Mistral-7b in several metrics. Future work could explore enhancing the pipeline with logical forms like lambda calculus for more analytical text generation, especially in complex domains like finance. Additionally, alternative intermediate representations, such as semantic graphs, may improve flexibility and handling of more intricate tabular formats. Expanding the dataset to new domains and incorporating variability in sentence structure could further enhance text quality, while also capturing trends and comparisons across rows or columns.

\section{Limitations}
The dataset used for our experiments contains tables from three specific domains. Subjectivity within the same type of table may be influenced by recurring semantics, which might not generalize well across more diverse datasets. Therefore, building a system capable of handling a wider variety of tables would benefit from a larger, more diverse dataset. Another limitation is the lack of established metrics to evaluate subjectivity in text generation tasks. Existing metrics such as BLEU, ROUGE, and METEOR focus on lexical overlap and semantic similarity but fail to capture nuanced aspects of subjectivity, such as sentiment, tone, and evaluative depth. This makes the evaluation of subjective text generation heavily reliant on human judgment, which can be subjective and inconsistent. Developing robust, automated metrics for subjectivity evaluation remains an open challenge and an important direction for future work.

\section{Ethics Statement}
The data used in this work were sourced from open-access platforms and did not contain any personal, restricted, or illegal information. None of the generated or annotated texts were designed to promote or disparage any team, entity, or individual.

\bibliographystyle{acl_natbib}
\bibliography{custom}

\input{appendix}

\end{document}

%% file: appendix.tex
\appendix

\section{Appendix}
\label{sec:appendix}

\subsection{Ta2TS Dataset Statistics}
\label{appendix:ta2ts}

The data distribution across the three domains of \emph{Finance, Weather and Sports} along with the average sizes of tables in Ta2TS dataset are highlighted in Table \ref{tab:datadis}.
\begin{table*}[ht]
    \centering
    \begin{tabular}{r c c c}
    \hline
        Genre & No. of Instances & Avg. no. of rows & Avg. no. of columns \\
        \hline
        \textbf{\large Finance} & & & \\
        Income statement & 500 & 11 & 5\\
        Balance Sheet & 499 & 11 & 5\\
        Cash flow & 499 & 11 & 5\\
        \hline
        \textbf{\large Weather forecast} & & & \\
        District-wise tables & 1031 & 5 & 10\\
        \hline
        \textbf{\large Sports} & & & \\
        Tournament Points Table & 409 & 9.4 & 7.2\\
        Series Form Table & 941 & 5 & 6 \\
        \hline
        \textbf{Total} & \textbf{3849} &   & \\
        \hline
    \end{tabular}
    \caption{\textbf{Genres in the Ta2TS Dataset}: The Ta2TS dataset comprises three primary genres, namely, financial tables, weather forecast tables and sports tables. In this table, the number of tables from each domain and different types are shown.}
    \label{tab:datadis}
\end{table*}

\subsection{Examples of Ta-G-T Generations}
\label{appexm}

\begin{figure*}[t]
    \centering
    \includegraphics[scale=0.55]{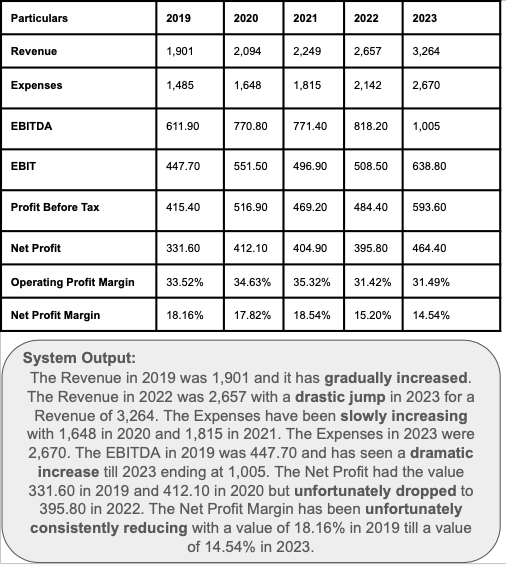}
    \caption{Subjective text description by Ta-G-T pipeline for a financial domain table from Ta2TS Dataset}
    \label{fig:finTable}
\end{figure*}

\begin{figure*}[t]
    \centering
    \includegraphics[scale=0.55]{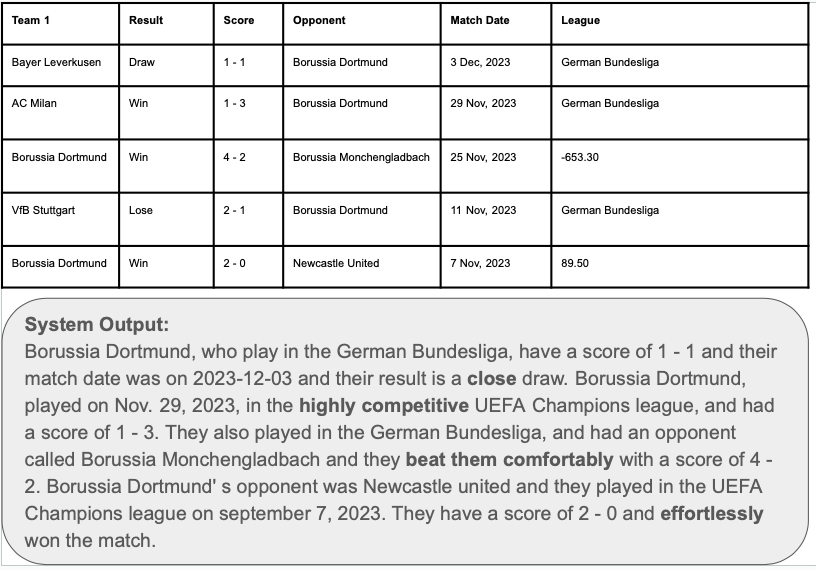}
    \caption{Subjective text description by Ta-G-T pipeline for a sports domain table from Ta2TS Dataset}
    \label{fig:sportTable}
\end{figure*}

Figure \ref{fig:finTable} shows an example of subjective text generated from a table in the Ta2TS Dataset. The table is from the financial domain and has information about the Income Statement of a company for 5 years from 2019 till 2023. The table covers key metrics like Revenue, Expenses, EBITDA, etc. The subjective text generated by the Ta-G-T system is given below. The subjective phrases are highlighted in bold and are used to emphasize the trends and insights obtained from the table.
Figure \ref{fig:sportTable} illustrates a subjective text generation example by the Ta-G-T pipeline for a sports domain table from the Ta2TS dataset. The table describes the performance of the football team called "Borussia Dortmund" for its previous five matches, and covers facts like the opponent team, match date, scores, etc. Phrases in bold are used to highlight the interpretative subjectiveness of the text.

\subsection{Human Evaluation Guidelines}
\label{appendix:humaneval}

To ensure consistency and reliability in our human evaluation process, we employed three evaluators who are postgraduate students in computer science and proficient in English. These evaluators were provided with a detailed set of guidelines to assess the generated text across four key dimensions: \textbf{Coherence, Coverage, Accuracy, and Subjectivity Capture}. Each evaluation was conducted on a \textbf{scale of 1 to 10}, where 1 represents the lowest quality and 10 represents the highest quality.

\subsection{Scoring Guidelines}

The evaluators followed the criteria outlined below to ensure consistency in the scoring process:

\subsubsection{Coherence}
Coherence measures the logical flow of the generated text, ensuring that sentences are well-connected and form a clear narrative. Evaluators were instructed to assess the fluency and readability of the text based on the following scale:

\begin{itemize}
    \item \textbf{Score 1-3}: The text is highly disjointed, with abrupt transitions, repetitive phrases, or lack of logical connections.
    \item \textbf{Score 4-6}: Some logical flow is present, but occasional breaks in fluency or awkward transitions make the text harder to follow.
    \item \textbf{Score 7-8}: The text is mostly coherent, with well-structured sentences and minimal fluency issues.
    \item \textbf{Score 9-10}: The text exhibits excellent coherence, with smooth transitions and a strong narrative structure.
\end{itemize}

\subsubsection{Coverage}
Coverage evaluates whether all essential information from the table is included in the generated text. Evaluators were specifically instructed to ignore inferred (dependent) information and focus only on independent facts.

\begin{itemize}
    \item \textbf{Score 1-3}: A significant portion of the necessary information is missing, making the text incomplete.
    \item \textbf{Score 4-6}: Some key details are omitted, but the text still conveys partial information from the table.
    \item \textbf{Score 7-8}: Most essential details are included, with only minor omissions.
    \item \textbf{Score 9-10}: The text comprehensively covers all necessary details from the table without unnecessary omissions or excessive repetition.
\end{itemize}

\subsubsection{Accuracy}
Accuracy measures whether the generated text correctly represents the objective facts from the table. Unlike coverage, accuracy does not penalize omissions but focuses on whether the included information is factually correct.

\begin{itemize}
    \item \textbf{Score 1-3}: The text contains multiple factual errors, misrepresentations, or contradictions with the table.
    \item \textbf{Score 4-6}: Some factual inconsistencies are present, though most of the included information remains accurate.
    \item \textbf{Score 7-8}: The text is mostly accurate, with only minor factual errors that do not significantly alter the meaning.
    \item \textbf{Score 9-10}: The text is fully accurate, correctly reflecting the tabular data without distortions or factual mistakes.
\end{itemize}

\subsubsection{Subjectivity Capture}
Subjectivity capture assesses the inclusion and appropriateness of subjective phrases that enhance the interpretability of tabular data.

\begin{itemize}
    \item \textbf{Score 1-3}: No subjective phrases are present, or those included are irrelevant, misleading, or inappropriate.
    \item \textbf{Score 4-6}: Some subjective elements are included, but they are either too generic or inconsistently applied.
    \item \textbf{Score 7-8}: The subjective phrases are mostly relevant and improve the interpretability of the data, with only minor inconsistencies.
    \item \textbf{Score 9-10}: The text successfully incorporates well-placed subjective elements that provide meaningful interpretations while maintaining factual correctness.
\end{itemize}

\subsection{Inter-Annotator Agreement}
\label{appendix:iaa}

\begin{table}[ht]
    \centering
    \begin{tabular}{c c}
        \hline
        \textbf{Metric} & \textbf{Fleiss' $\kappa$} \\
        \hline
        \textbf{Coh.} & 0.78 \\
        \textbf{Cov.} & 0.75 \\
        \textbf{Acc.} & 0.82 \\
        \textbf{Sub.} & 0.68 \\
        \hline
    \end{tabular}
    \caption{\textbf{Inter-Annotator Agreement for Ta-G-T}: Fleiss’ $\kappa$ values for agreement across all three evaluators for Coherence (Coh.), Coverage (Cov.), Accuracy (Acc.) and Subjectivity Capture (Sub.).}
    \label{tab:iaa}
\end{table}

To assess the reliability of human evaluations, we computed Fleiss’ $\kappa$, which measures agreement across the three evaluators. Table~\ref{tab:iaa} shows that accuracy has the highest agreement ($\kappa = 0.82$), while subjectivity capture has the lowest ($\kappa = 0.68$), reflecting the inherent variability in subjective interpretation. Overall, the agreement levels indicate substantial consistency among evaluators.